# Transforming Dental Diagnostics with Artificial Intelligence: Advanced Integration of ChatGPT and Large Language Models for Patient Care

Masoumeh Farhadi Nia[1], Mohsen Ahmadi[2, *], Elyas Irankhah[3]

1- Department of Electrical and Computer Engineering, University of Massachusetts Lowell, Lowell, MA, USA

2- Department of Electrical Engineering and Computer Science, Florida Atlantic University, FL, USA

3- Department of Mechanical Engineering, University of Massachusetts Lowell, Lowell, MA, USA

*Corresponding author: mahmadi2021@fau.edu

## Abstract

Artificial intelligence has dramatically reshaped our interaction with digital technologies, ushering in an era where advancements in AI algorithms and Large Language Models (LLMs) have natural language processing (NLP) systems like ChatGPT. This study delves into the impact of cutting-edge LLMs, notably OpenAI's ChatGPT, on medical diagnostics, with a keen focus on the dental sector. Leveraging publicly accessible datasets, these models augment the diagnostic capabilities of medical professionals, streamline communication between patients and healthcare providers, and enhance the efficiency of clinical procedures. The advent of ChatGPT-4 is poised to make substantial inroads into dental practices, especially in the realm of oral surgery. This paper sheds light on the current landscape and explores potential future research directions in the burgeoning field of LLMs, offering valuable insights for both practitioners and developers. Furthermore, it critically assesses the broad implications and challenges within various sectors, including academia and healthcare, thus mapping out an overview of AI's role in transforming dental diagnostics for enhanced patient care.

## 1- Introduction

Artificial intelligence is expected to revolutionize sectors such as healthcare and dentistry by presenting new approaches to various clinical problems. This would enhance the productivity of medical practitioners. The ChatGPT model has been developed by OpenAI using the GPT framework (Generative Pretrained Transformer), a deep learning model that is utilized in natural language processing (NLP) [1]. The model can produce text that is like human writing and can interact with users via chat platforms. With ChatGPT's generative AI, original content can be generated in real-time discussions [2,3]. A variety of artificial intelligence models are employed to provide conversational responses to inquiries based on extensive text data. ChatGPT is designed to remember both user input and its own responses within a conversation, allowing it to build upon previous responses as new inquiries are received [4]. The effectiveness of ChatGPT has been considered in several healthcare domains, including its capacity to diagnose dental problems. It has been demonstrated that it provides accurate differential diagnoses in a substantial number of cases, compared favorably with the diagnostic abilities of healthcare professionals [5].

In the realm of dentistry, research on ChatGPT's effectiveness remains limited. There has been research on ChatGPT's capability to produce scientific content within oral and maxillofacial surgery, as well as its accuracy and dependability in providing concise clinical responses in endodontics [6-8]. In addition, its proficiency has been assessed in various scenarios, including board-style dental knowledge quizzes, and answering questions related to scientific or research writing. Health care and dentistry will benefit greatly from the deployment of ChatGPT because it will increase patient autonomy, enhance the efficiency and safety of services, boost sustainability, and broaden the range and quality of care, ultimately empowering patients [9]. Artificial intelligence finds numerous applications in prosthodontics, such as implant-supported prosthetics, computer-aided design (CAD), maxillofacial prosthetics, computer-aided manufacturing (CAM), and both fixed and removable prosthetics [10]. Research on ChatGPT's application within medical and dental disciplines lacks extensive systematic reviews and meta-analyses. The utility of ChatGPT has been examined in numerous healthcare contexts including dental, However, these studies often have limited scope and fail to provide insights into the potential benefits and drawbacks of integrating ChatGPT into these fields., often consisting of only literature reviews and editorials [11-13]. It is anticipated that ChatGPT will enhance the creation of precise and documentation, facilitating collaboration and knowledge exchange among oral medicine practitioners across clinics, hospitals, and departments [12]. Oral medicine professionals can utilize ChatGPT's text-generation features to input patient data, clinical observations, and treatment strategies, enabling them to create detailed reports. This not only enhances the quality of documentation but also facilitates effective communication among healthcare providers. Moreover, oral medicine specialists can utilize ChatGPT for virtual patient interactions, offering advice, responding to inquiries, and discussing treatment options. It also assists in the decision-making process in cases involving complex oral medicine. The primary objective of this review was to provide a detailed, evidence-based evaluation of ChatGPT's potential as a resource for medical and dental research, with the goal of directing future investigations and informing clinical practice.

## 2- AI and Data Analysis in Dentistry

Within the medical community, ChatGPT is gaining recognition for its influence on both the healthcare system and medical field. It provides support as a complementary resource for diagnosis and decision-making across a variety of medical disciplines [14]. Although the accuracy of its outputs and the potential for reinforcing biased diagnoses have prompted discussions on the necessity of human oversight when using this technology, ChatGPT can also be utilized in assessing disease risks and outcomes, advancing drug development, enhancing biomedical research, and transforming healthcare practice in significant ways [15-17]. Artificial intelligence has made remarkable progress in the field of digital health, particularly in dentistry. In addition to diagnosing dental issues through the analysis of imaging, it also aids in the planning of medical operations [18]. The capabilities of artificial intelligence extend to the analysis of auditory data to improve understanding of oral functions, thereby supporting dental education. The potential applications of AI in endodontics are demonstrated in Figure 1.

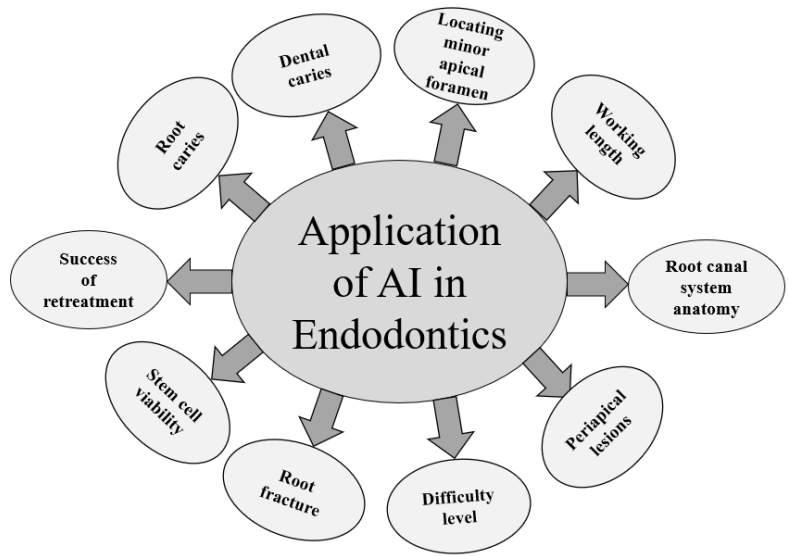

**Figure 1**: The Multifaceted Effects of Artificial Intelligence on Endodontic Practice [19]

As evidenced by a bibliometric analysis showing an increase in publications on the subject from a single article in 2000 to 120 in 2023, with a peak of 208 articles in 2022, AI's applications within dentistry are on the rise [20]. The use of neural networks in diagnosing endodontic problems has been proven to be as accurate as that of dental professionals, offering benefits to nonspecialists and those new to the field [21]. Studies have shown the wide-ranging potential of artificial intelligence in endodontics, emphasizing how it can enhance diagnostic precision, treatment planning, and educational outcomes. Additional AI applications are further described in this research and are briefly summarized in Figure 2.

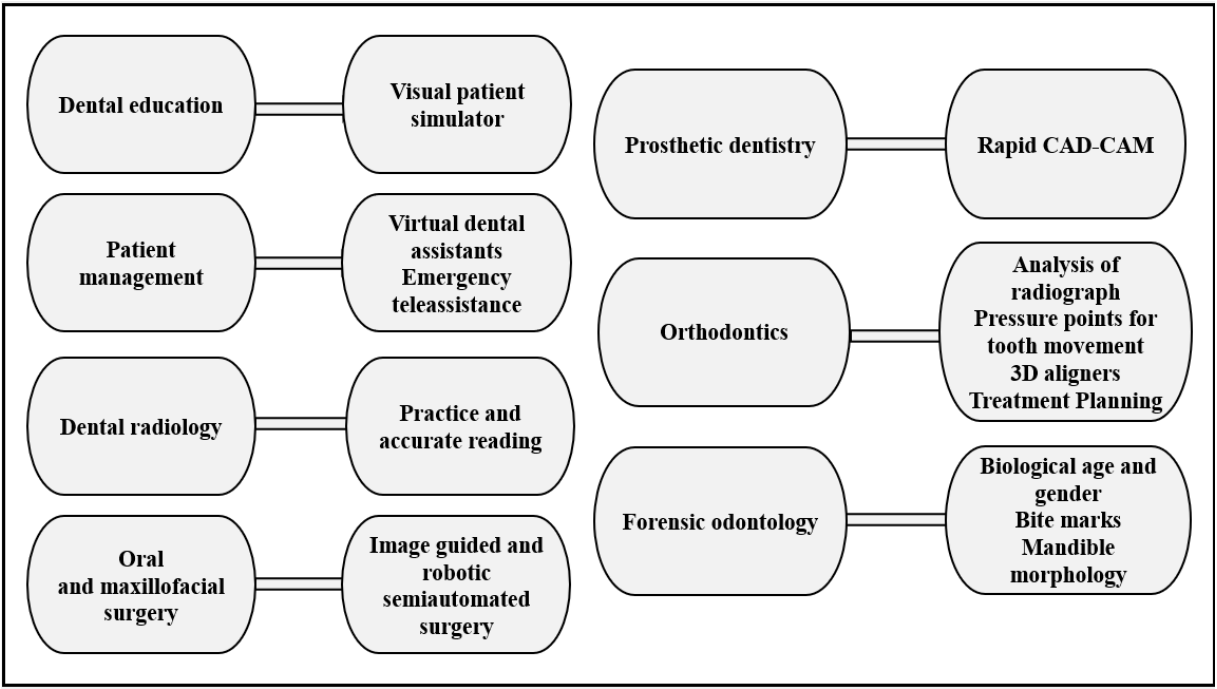

**Figure 2**: Diverse Applications of AI in Dental Specialties

Mohan et al. [22] assert that despite concerns regarding patient privacy and potential misinterpretations, The integration of AI-driven care systems in dentistry will improve patient care, encourage innovative research, and require collaboration across various professional fields in the future. Additionally, OpenAI's Generative Pre-Trained Transformer, GPT-4, represents an advancement in AI language technologies, impacting dentistry and healthcare profoundly [23]. Mohamed M. Meghil et al. [24] present a systematic review adhering to PRISMA standards and the Cochrane Handbook, analyzing the evolution and integration of artificial intelligence (AI) in various dental specialties. Their research encompasses 28 studies from distinct fields such as pathology, maxillofacial surgery, and orthodontics, among others, sourced from PubMed and Web of Science. These studies were evaluated for methodological quality and bias with the PROBAST tool, highlighting AI's role in clinical data analysis across dentistry. The findings underline the substantial datasets available for training and testing, indicating AI's promise for improving patient outcomes, diagnostic processes, and treatment planning. However, the authors advocate for additional research, specifically randomized clinical trials, to solidify AI's efficacy in dental practice, aiming for data-driven, superior dental care that revolutionizes treatment delivery.

The review by [25] delves into AI's journey from its conceptual inception in 1950 to its practical applications in dentistry today, fueled by advancements in data analytics, algorithm development, and computational power over the last two decades. It examines AI's impact across several dental specialties, notably in diagnostic uses involving optical and radiographic imaging. The paper also addresses the challenges AI faces, such as data scarcity, standardization, and the computational demands of 3D data management. Furthermore, it explores the potential synergy between evidence-based dentistry (EBD) and machine learning (ML), suggesting ML can enhance dental practitioners' clinical decision-making and efficiency. Marta Revilla-León et al. [26] conduct a systematic review on AI's application in implant dentistry, focusing on design optimization, implant type identification, and success prediction. Through a precise search of five databases up to February 21, 2021, plus manual searches, seventeen studies were selected and evaluated utilizing the Joanna Briggs Institute Critical Appraisal Checklist. These studies, which include imaging for implant identification and the use of AI for predicting implant success and optimizing design, demonstrate promising accuracy and potential for enhancing implant dentistry. The research indicates accuracy in implant recognition and suggests areas for improvement in success prediction models and design optimization through AI. Another study [27] examines how AI could transform dental imaging, conducting a systematic review and meta-analysis. It evaluates the influence of AI, particularly utilizing convolutional neural networks and deep learning algorithms, on the precision and speed of dental imaging tasks. The analysis incorporates nine studies showcasing AI's ability to improve diagnostic accuracy in dental imaging, indicating significant advantages for patient care. Despite these advancements, the study acknowledges the need for more clinical trials to address limitations and fully leverage AI in dental imaging.

In [28] explores AI's application in dentistry within low- and middle-income countries (LMICs), identifying a research gap compared to global trends. Through an extensive search across Scopus, PubMed, and EBSCO Dentistry & Oral Sciences Source, 25 relevant studies were selected from 1587 publications, mainly focusing on orthodontics. The review highlights the lack of research on cost-utility and patient experiences in LMICs, underscoring the need for broader, more inclusive studies to understand and utilize AI for improving dental care in these regions.

The modeling of language has evolved from simple n-gram models to more sophisticated frameworks such as the GPT series, Bidirectional Encoder Representations from Transformers (BERT), and Embeddings from Language Models (ELMo). ChatGPT is a crucial development in the progression of large language models (LLMs), demonstrating exceptional conversational abilities because of extensive pretraining across a wide range of domains. This includes multimodal LLMs, extensive pretraining in both audio-language and vision-language, positioning LLMs as versatile and adaptive tools. The study [23] examines the incorporation of LLMs in dentistry through two primary methods: cross-modal and automated dental diagnosis. These approaches utilize LLMs' developed NLP and ability to assess varied data sources, facilitated by cross-modal encoders. Automated strategies include text mining for medical records, natural language reasoning for treatment planning, and NLP for medical documentation. Cross-modal dental diagnosis employs vision-language and audio-language techniques for visual question answering, visual grounding, and audio data diagnostics. The research also addresses AI in dentistry's challenges, such as data quality, model bias, privacy concerns, and computational limitations. It suggests human oversight to correct LLMs' potential inaccuracies, proposes neural-symbolic models to mitigate bias, emphasizes strict data privacy protocols, and recommends sparse expert models to address computational issues. Chat systems could pre-educate patients, leading to more effective practitioner consultations. Evaluating these systems' accuracy in answering medical questions is highlighted as a critical area of dental research.

## 2-1- ChatGPT and Electronic Dental Records (EDRs)

To enhance data management, improve diagnostic precision, and customize patient care in the dental field, ChatGPT needs to be integrated with Electronic Dental Records (EDRs). A study conducted by Patel et al. [30] examined the utility of Electronic Dental Records (EDRs) for classifying patients' smoking status, including the intensity of smoking habits. Three machine learning classifiers were employed, random forest, support vector machine, and naive Bayes. It was found that the support vector machine was most effective at accurately classifying smoking status at various intensities. The study revealed that EDRs have the potential to provide detailed insights into patients' smoking behavior, an aspect often overlooked by Electronic Health Records (EHRs). The use of EDRs is limited to secondary purposes, such as conducting research or analyzing treatment outcomes. In addition, [31] discusses both the challenges and opportunities associated with leveraging EDRs for dental research. There is a prevalent use of unstructured text in EDRs, which makes their use in research or outcome analysis difficult. While NLP techniques are acknowledged for their ability to extract detailed information from clinical notes in EDRs, the paper highlights obstacles to their widespread adoption. Among these challenges are the lack of applications of NLP in primary oral healthcare, the difficulty of generalizing findings due to dentistry's unique vocabularies, and the difficulties associated with cross-referencing results and validating them. An overview of ChatGPT and EDRs is shown in Table 1.

**Table 1:** An overview of ChatGPT and Electronic Dental Records (EDRs)

| Researchers | Techniques | Data Quantity | LLM Categories | Inquiry Technique | Evaluation Criteria |
|---|---|---|---|---|---|
| Mago & Sharma (2023) [38] | Interrogating an LLM about imaging features, oral anatomy, and disease | Set of 80 queries | Conversational Model GPT-3 | Unprompted | 4-tier modified preference scale |
| Chuang et al. (2023)[130] | Use of GPT-J for prompt generation in NER models for diagnosis extraction from electronic dental records | Dataset of 5,495 eligible patients, resulting in 8,125 clinical notes for periodontal diagnosis extraction | GPT-J, RoBERTa with spaCy package | Direct testing and seed generation for feeding NER models | Performance measured by F1 score, highlighting the importance of seed quality. Consistent performance across settings with F1 scores between 0.92-0.97 after RoBERTa training. |
| Doshi et al. (2023) [131] | Refinement of imaging summaries | 254 imaging summaries | Conversational Models GPT-3.5, GPT-4, Bing, and Bard | Varied query structures | Comprehensibility metrics |
| Mykhalko et al. (2023) [132] | Evaluating diagnostic capabilities using ChatGPT-3.5 with various chat setups | 50 clinical cases | ChatGPT-3.5 | Three-phase experiment with different information and prompt setups | Diagnostic accuracy percentages across phases (66%, 70.59%, and 46%), highlighting the strength of ChatGPT in structured scenarios and the importance of prompt engineering. |
| Lai et al. (2023) [133] | Evaluating ChatGPT-4's performance on the United Kingdom Medical Licensing Assessment (UKMLA) | 191 SBA questions from UKMLA | ChatGPT-4 | Three attempts over three weeks with structured SBA questions | Average performance score of 76.3% across three attempts, with analysis on areas of both strength and weakness in specific medical fields. Consistency in correct and incorrect responses evaluated. |
| Jeblick et al. (2023) [134] | Condensing imaging narratives | Trio of imaging narratives | Conversational Model GPT | Singular query structure | 5-tier preference scale |
| Lyu et al. (2023) [135] | Converting imaging narratives into layman's terms | 62 thoracic CT and 76 cranial MRI summaries | Conversational Model GPT-4 | Multiple query structures | 5-tier preference scale |
| Waters et al. (2023) [136] | Practical applications and suggestions for interacting with LLMs like ChatGPT in radiation oncology | Not specified | ChatGPT | Guide on how to interact with LLMs in clinical and administrative tasks | Highlights the potential uses, limitations, and ethical considerations of LLMs in radiation oncology, emphasizing the importance of human review. |
| Chiesa-Estomba et al. (2023) [72] | Evaluating Chat-GPT as an aiding tool for sialendoscopy clinical decision-making and patient information support. | Not specified | ChatGPT | Prospective, cross-sectional study comparing Chat-GPT with expert sialendoscopists | The study compared Chat-GPT and expert sialendoscopists' agreement on salivary gland disorder management (Chat-GPT: 3.4, experts: 4.1) and evaluated therapeutic alternatives suggested by both." |

The study [33] explored ChatGPT's proficiency in answering frequently asked questions (FAQs) regarding fluoride based on the American Dental Association (ADA) guidelines. Using structured methodology, ChatGPT was asked a series of eight specific fluoride-related questions on two separate occasions, May 8th and May 16th, 2023. The recorded responses from ChatGPT were compared with those from the ADA's website. ChatGPT's responses and those provided by the American Dental Association were analyzed primarily qualitatively to identify similarities in wording and context. In this assessment, ChatGPT's answers were compared with the official ADA answers to assess their consistency over the week-long period. While ChatGPT's responses were consistent over time, they were more detailed and scientific than those of the ADA, even though both fundamentally conveyed the same information about the role of fluoride in dental health. In addition to radical dental information dissemination, these studies present promising applications in radiology for more efficient and accurate medical image diagnoses, which could result in improved patient care and lower healthcare costs.

## 2-2- An evaluation of ChatGPT performance in prosthodontics

Employing AI technology in clinical applications has enhanced patient results, made processes more efficient, and reduced costs. As a result of AI's tremendous success in medical practice, it has demonstrated its potential to transform healthcare services by segmenting brain tumors, aiding clinical decision-making through epidemiological forecasts, and performing complex surgical and rehabilitation procedures [34-36]. Convolutional neural networks have demonstrated improved accuracy in identifying and classifying maxillofacial fractures in dentistry. It is noteworthy that previous studies have examined the precision of ChatGPT-generated content in the fields of neck and head surgery, and oral and maxillofacial surgery [37]. GPT-3 was found to accurately identify anatomical points in oral and maxillofacial radiography with 100% accuracy, although Mago et al. [38] cautioned against relying solely on GPT-3 due to its occasional lack of detail and potential inaccuracies. In addition, Vaira et al. [39] carried out a multi-site study in August 2023 to evaluate the accuracy of the answers produced by ChatGPT regarding head and neck oral maxillofacial surgery. They found that 87.2% of the responses were completely or almost completely correct, and 73% were comprehensive. Despite these high percentages, the researchers determined that AI is not a dependable tool for decision-making in cases related to head-neck surgery. According to Vaira et al., their results exceeded our own, which calculated accuracy and completeness percentages only based on responses that were correct in every respect. It is of crucial importance to assess the accuracy and consistency of ChatGPT in prosthodontics. An assessment of the accuracy of ChatGPT would involve posing a series of prosthodontics-related questions and analyzing the responses considering current dental practices and literature. As part of the accuracy measurement, ChatGPT's answers are compared to established knowledge and guidelines in prosthodontics, whereas the consistency (or repeatability) measurement examines whether ChatGPT provides the same or very similar answers when asked the same questions several times or under slightly different phrasings. At present, there is a lack of scientific literature focusing on the application of ChatGPT to interceptive orthodontics.

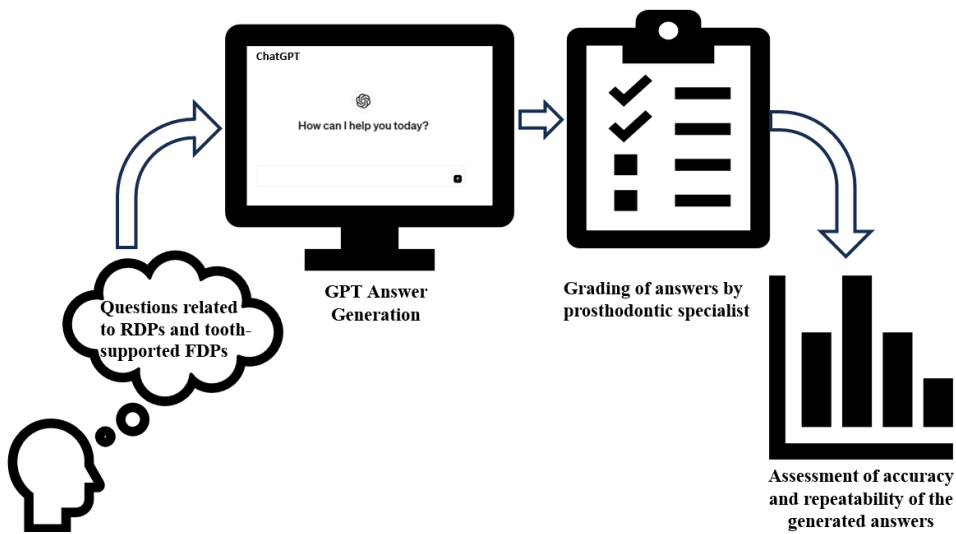

**Figure 3:** The approach employed to assess the accuracy and consistency of responses generated by ChatGPT to questions regarding RDPs and tooth supported FDPs.

Figure 3 presents a flowchart outlining the process for assessing the accuracy and consistency of responses generated by ChatGPT concerning questions related to Removable Dental Prostheses (RDPs) and tooth-supported Fixed Dental Prostheses (FDPs). It begins with the formulation of questions regarding to tooth supported FDPs and RDPs. The questions are then entered into the ChatGPT interface, which begins the process of generating GPT answers. A prosthodontic specialist grades the responses obtained from ChatGPT based on their accuracy and relevance [34,35]. As a final step in the methodology, the graded answers are statistically analyzed to determine both the accuracy of the ChatGPT answer and the repeatability of the answers across multiple interactions. In the field of orthodontics, Subramanian et al. [40] demonstrate AI's potential to simplify cephalometric tracing for everyday clinical use and to conduct extensive data analysis for research. In a study by Tanaka et al. [41], they found that ChatGPT was effective in delivering informative responses on topics related to orthodontics, such as digital imaging, clear aligners, and temporary anchorage devices. Tanaka et al.'s study is similar to the current study but differs in those ten orthodontists with specialized training from various Italian orthodontic schools evaluated the patients instead of five general orthodontists as in Tanaka et al.'s study. Duran et al. [42] acknowledge that while ChatGPT can provide reliable information on cleft lip and palate, its complex responses need professional verification. The AI tool, aware of its limitations, advises consulting a specialist for orthodontic decisions. Despite its usefulness, ChatGPT cannot replace the essential in-person aspects of dental care. However, it could streamline administrative tasks in dental practices, such as managing insurance claims and supporting dental telemedicine. Gonzales et al. [43] suggest that AI can match expert human observers in determining cervical vertebral maturity stages, potentially increasing diagnostic precision, and improving orthodontists' efficiency. The authors of this study hold a different view, noting that the orthodontists involved did not find the ChatGPT responses completely accurate. For both open-ended and clinical case questions, accuracy scored 4.9 out of 6, while completeness scored 54.3%. Furthermore, Ahmed et al. [44] have shown that AI can develop precise caries detection models, aiding clinical decision-making, and improving patient care. AI has not yet substantially improved the ability to provide answers to open-ended questions, diagnoses, or treatment planning for clinical cases, despite X-ray analysis being a static process.

## 3- AI-based systems developed and medical training

AI in healthcare is frequently spearheaded by computer scientists without medical training, leading to an emphasis on problem-solving rather than the holistic, expertise-driven care offered by clinicians [45]. However, despite AI's advantages in dentistry, such as speed, accuracy, and standardization of procedures, dental professionals are hesitant to adopt such technologies, partly due to steep learning curves, high implementation costs, extensive data, and training requirements. [46]. Further, AI applications are not always directly applicable to clinical practice. AI has demonstrated promise in medical imaging, such as in identifying and diagnosing COVID-19 based on CT scans [47], and has made strides in other fields, such as facial recognition. The transition of artificial intelligence from a perceived threat to a valuable tool is a reflection of its growing influence in society, healthcare, and dentistry [48-50]. Dentists are cautioned, however, against over-relying on artificial intelligence for diagnosis and treatment. An emerging tool in dentistry that mimics human-like text responses is ChatGPT [51]. Through machine learning (ML) and deep learning (DL), it is aimed at improving computer communication, language processing, and responsiveness. There are several potential benefits of ChatGPT across a variety of sectors, including education, where it can assist students in learning and understanding complex topics, and it has potential as a pedagogical tool in biomedical sciences [51,52]. Specifically for patients who are pre- and post-surgical, ChatGPT provides valuable information and education, answering medical questions and setting realistic expectations for surgical outcomes. ChatGPT can enhance patient empowerment, service efficiency, safety, and access to quality care in healthcare and dentistry, as well as support patient education and decision-making. Lee et al. [53] delves into the transformative possibilities of large language models, such as ChatGPT, in medical education, emphasizing their potential to enhance learning through interactive simulations and act as virtual teaching assistants. Lee [54] suggests that these models could enhance student engagement and learning outcomes, though calls for more research to substantiate these benefits. The discussion also acknowledges the possible drawbacks and ethical points in regards with ChatGPT, emphasizing the need for medical educators to adapt to technological advancements in teaching strategies, curriculum development, and evaluation methods. Lee concludes that ongoing research and thorough assessment are important for the effective integration of AI tools in medical education to achieve educational goals.

In [54] studied the necessity for updates in medical school curricula to align with rapid AI advancements, particularly highlighting the role of ChatGPT. The editorial outlines a framework for ChatGPT's integration into medical education, proposing immediate actions like enhancing digital literacy and long-term strategies such as prioritizing the human aspect of healthcare. These suggestions aim to empower medical students with the proficiency to navigate the AI-influenced landscape of future healthcare. A team of 29 researchers in [55] conducts a study to assess ChatGPT's efficacy in responding medical questions. The study involved 33 doctors from 17 specialties, who generated 284 questions answered by ChatGPT. These responses were evaluated for accuracy and completeness, showing promising results that suggest ChatGPT could serve as a reliable medical knowledge resource. However, the study also notes variability in performance based on question complexity, indicating areas for further improvement and research. The cross-sectional study in [56] investigates perceptions among recent medical graduates regarding the use of ChatGPT and AI in their education and future careers. A survey conducted at an international academic medical center revealed cautious use of ChatGPT among students, with interest in leveraging AI for learning, research, and exam preparation during residency. The findings suggest a gender difference in attitudes towards AI's potential benefits in healthcare, with positive views linked to prior AI experience. The study underscores the need for formal AI education policies to prepare students for AI-integrated medicine. [57] delves into the advancements brought about by ChatGPT in artificial intelligence, providing a literature review on its capabilities, training processes, and potential impacts on technology and understanding. This review highlights ChatGPT's innovative approaches, including NLP, Supervised Learning, and Reinforcement Learning, offering a critical examination of the challenges to be addressed for maximizing its utility and innovation in various fields.

It is important to note that ChatGPT, like any other emerging technology, has its limitations. The system is operated by a vast neural network that is extremely computationally intensive and consumes a large amount of memory, making it prohibitively expensive for smaller medical practices with limited budgets [58]. The model's inability to incorporate external data sources can negatively affect its accuracy, which is a critical factor in the medical field. Further, ChatGPT sometimes fails to cite references correctly or at all, and its responses may differ based on the context of the conversation, leading to convincingly incorrect responses [58-61]. It is for these reasons that ChatGPT must properly express uncertainty to avoid misdirection. The advent of artificial intelligence technologies in dentistry, referred to as "Dentronics," heralds a new phase of transformation with potential benefits, including improved reliability, reproducibility, precision, and efficiency. Additionally, Dentronics could enhance the understanding of disease mechanisms, risk assessments, diagnostic processes, and prognoses to improve patient outcomes [61]. The integration of AI into dentistry continues to grow, however, it is unlikely to completely replace dentists, since the profession demands more than just disease diagnosis. It also involves integrating clinical observations and providing patient care. In spite of this, dental professionals would benefit from gaining a solid understanding of AI principles and methodologies as their field continues to evolve [62,63].

## 4- AI for dental pathology

Medical and scientific landscapes have been transformed by artificial intelligence, which has been incorporated into a variety of aspects of diagnosis, therapy, and patient care [64]. The effectiveness of AI chatbots in providing accurate and trustworthy information is vital for their adoption in clinical decision-making and patient care. These tools can facilitate discussions that may enhance diagnostic and treatment guidelines. In the case of the maxillary sinuses, for instance, AI can be instrumental in identifying pathologies that may not be apparent on extraoral radiographs, thus reducing diagnostic errors, especially for less experienced dentists. Using Water's view radiographs of the maxillary sinus, Kim et al [65]. demonstrated superior sensitivity and specificity compared to radiologists. Similarly, AI has been shown to effectively identify conditions such as mucosal hypertrophy and retention cysts that are sometimes overlooked by radiologists. Another study proposed a CNN model to assist in distinguishing these conditions from CBCT images by Kuwana et al [66]. Early detection is vital in the case of oral cancer, which ranks as the sixth most prevalent type of cancer. As a result of AI, early detection may be enhanced, thereby potentially reducing mortality and morbidity associated with the disease. By using laser-induced autofluorescence spectra, Nayak et al. [67] were able to distinguish normal from premalignant and malignant tissues with a high level of accuracy, specificity, and sensitivity, suggesting its application in real-time clinical setting. The development of AI models capable of predicting cancer recurrence and occurrence has led to more accurate diagnostic tools and improved patient care as a result of ongoing research in oral cancer. Applications of artificial intelligence in clinical decision support can assist in screening oral mucosal lesions, classifying suspicious changes, diagnosing tissues, estimating lymph node involvement, examining gene expression, and profiling microbiota. In [68], researchers explore the capability of artificial intelligence, particularly ChatGPT, in recognizing oral and maxillofacial disorders. They highlight AI's potential to process large volumes of patient information, aiding dental practitioners in diagnosis and treatment planning. With advancements like GPT-4.0, which can interpret information from images, ChatGPT is poised to enhance diagnostic accuracy and treatment approaches. While it can direct patients towards timely professional consultations, offering a more reliable health advice source than conventional search engines, the authors stress that ChatGPT should not replace professional consultation and underscore the importance of further clinical validation.

A systematic review in [69] focused on ChatGPT's role in diagnostic human pathology reveals the capabilities of the model as a supportive equipment for pathologists by providing extensive scientific information. Despite identifying a limited number of studies, a "question session" with ChatGPT covering various pathologies suggests its usefulness in diagnostics. However, the review also highlights challenges, including issues with the quality of training data and occurrences of AI-generated "hallucinations". It underscores the role of AI as a support rather than a decision-maker, emphasizing the necessity for improvements in its application. Concerns regarding the accuracy of BARD and ChatGPT in generating scientific references, especially in oral pathology, are addressed in [70]. The comparison of references from both chatbots against reputable sources reveals a low accuracy rate, with ChatGPT achieving only a 10% accuracy rate. The study highlights the unreliability of these AI methods in producing accurate bibliographic information and calls for improvements, including integrating reliable scientific databases into chatbot systems to enhance reference accuracy. A review in [71] examines ChatGPT's impact on dental research, showcasing how its transformer-based architecture aids in the efficient processing and synthesis of textual data for systematic studies and literature reviews. While highlighting the model's potential to streamline research activities, it also points out potential drawbacks like understanding specific dental issues and ethical concerns related to bias in training datasets. The review emphasizes the significance of integrating AI's capabilities with human expertise to achieve well-rounded research approaches.

**Table 2**: Integration and Applications of Digital Dentistry in Modern Dental Practices

| Time Period | Application | Field | Critical Technologies |
|---|---|---|---|
| Past | Digitized dental impressions | Dental reconstructions | Computer-Aided Design/Manufacturing, Intra-oral digital scanning |
| Past | Assisted computer navigation for dental implants | Implantology | Cone Beam Computed Tomography (CBCT), Computer-Aided Design/Manufacturing |
| Present | Three-dimensional dental replicas | Prosthetic dentistry, Orthodontic treatment | Three-dimensional fabrication, Computer-Aided Design/Manufacturing |
| Present | Machine learning detection of tooth decay | Diagnostic Procedures | Machine learning, digital image processing |
| Present | Virtual reality for dental education | Educational tools for patients | Virtual reality, Three-dimensional visualization |
| Future | Machine intelligence for treatment strategizing | Orthodontic planning, Maxillofacial surgery | Machine intelligence, digital image interpretation |
| Future | Biocompatible dental components produced by additive manufacturing | Dental implantology | Additive manufacturing, Computer-Aided Design/Manufacturing, Bioengineering materials |
| Future | Remote dental consultations via digital communication | Remote dental care | Telecommunication, Digital communication platforms, diagnostic imaging technology |
| Future | Tailored dental prosthodontics | Prosthodontics | Digital dental mapping, Computer-Aided Design/Manufacturing, Three-dimensional fabrication |

| Future | Utilization of advanced language models in dentistry | Dental diagnosis automation and multimodal diagnostic evaluation | Advanced linguistic computational models, specifically ChatGPT |
|---|---|---|---|
| Future | Customized conversational assistance | Patient Consultation | Conversational AI, Chatbot technology |
| Future /present | Preparing prospective dental professionals | Dental undergraduate education | Augmented Reality (AR), Virtual Reality (VR) |
| Present | Advancement and evaluation of color matching in dental restorations | Dental restoration color matching | Traditional methods versus spectrophotometric benchmarking |
| Present | Transition to digital radiography | Routine dental operations | Enhancement of image luminance and contrast, specialized image processing algorithms, sensor-agnostic technologies |

In a study [21] conducted in Japan, deep learning was employed to detect fatty degeneration in salivary gland parenchyma, indicating Sjogren's syndrome. The study utilized 500 CT scans, with 400 for training and 100 for testing. The diagnostic performance of the DL system was on par with experienced radiologists and surpassed that of less experienced counterparts. As shown in Table 2, the followings are instances of the integration and application of digital dentistry in today's dental practices. Diagnosing salivary gland tumors is challenging due to their infrequent incidence and morphological similarities. Machine learning has been employed to discern malignant salivary gland tumors based on their cytological features. In a study by Chiesa-Estomba and colleagues [72], a comprehensive approach integrating clinical, radiological, histological, and cytological data was utilized to anticipate facial nerve dysfunction in patients undergoing surgical treatment for salivary gland tumors with potential posterior nerve injury. Artificial intelligence stands as a valuable diagnostic tool, offering the capability to predict facial nerve injury and providing proactive awareness for surgeons and patients about potential complications.

## 5- AI in Clinical Dental Practice and Surgery

The incorporation of Artificial Intelligence (AI) into clinical dental practice and surgery represents a paradigm shift, promising to refine diagnostic capabilities, streamline surgical precision, and revolutionize patient care. In a study [73], researchers evaluated ChatGPT utility in dental education and potential as a clinical decision support system. Gathering insights from 27 specialists in nine dental fields and two institutional heads, the study assessed ChatGPT's responses to 243 dental questions using Likert scales for accuracy, completeness, and relevance. Oral medicine and radiology received the highest ratings, indicating a general precision in ChatGPT's answers. While the results underscore

ChatGPT's efficacy in information retrieval and handling straightforward queries, they also caution against replacing professional consultations with AI advice, advocating for further evaluations before ChatGPT's clinical application. In [74] explores ChatGPT's role in enhancing otolaryngology surgical training, suggesting ways to leverage the AI model for educational gains, especially during the reduced learning opportunities caused by the COVID-19 pandemic. The authors, drawing from their experience, discuss ChatGPT's potential to mitigate educational challenges in otolaryngology, particularly in improving communication skills and providing tailored patient responses. They conclude that despite ChatGPT's inability to substitute traditional mentorship or clinical experience, it represents a valuable educational tool for otolaryngology trainees.

A study [75] assesses ChatGPT's performance in cases related to laryngology and head and neck surgery cases, comparing its recommendations against those of board-certified otolaryngologists. ChatGPT tended to suggest more tests than necessary, achieving consensus on common tests but missing key examinations like PET-CT scans. However, it accurately provided primary diagnoses and treatment options, underscoring its utility as a supplementary resource in diagnosis and treatment planning despite certain limitations. Thorat et al. [76] discuss Chatbot GPT's transformative impact on undergraduate dental education. The technology, particularly through its GPT-3 architecture, supports personalized learning, evidence-based assessment, and integration into virtual simulations, among other benefits. The paper highlights Chatbot GPT's role in enhancing learning diversity, offering risk-free clinical practice through simulations, and providing 24/7 multilingual support, positioning it as an indispensable tool in modern dental education. Qu et al. [77] conducted a cross-sectional survey in otolaryngology to evaluate ChatGPT 4.0's diagnostic and management capabilities. Using 20 clinical vignettes, the study found high agreement levels with ChatGPT's differential diagnoses and treatment plans among attending physicians, despite a noted decrease in management quality with reduced diagnostic accuracy. This indicates the potential of AI models like ChatGPT to refine patient management and medical diagnoses in otolaryngology, though further research is necessary to address its limitations with complex clinical data.

According to Saravanan's study [78], it is explored whether AI, and specifically ChatGPT, is a beneficial innovation or a complicated challenge in the realm of oral and maxillofacial surgery. AI is evaluated considering its integration into surgical procedures, its impact on patient care, and its balance with traditional surgical practices. The research examines the potential benefits that artificial intelligence may have in enhancing patient communication, surgical precision, and educational aspects of oral and maxillofacial surgery. The study provides a holistic perspective on AI's role in dentistry, highlighting both the benefits and challenges of these technological developments, contributing to a better understanding of AI's future direction in dental surgery. Furthermore, [79] extends the scope of this study to include AI's application in general clinical dental practice and surgery. As a result, it explores the application of specific AI tools in dental procedures, such as computer vision, machine learning algorithms, and transforming aspects of dental diagnostics, treatment formulation, and patient care. This study seeks to examine the contribution of artificial intelligence to personalized treatment approaches, predictive analytics for disease prevention, and the streamlining of routine tasks in dental practices, all to improve precision and operational efficiency. It is expected that This research will offer valuable perspectives on the transformative capacity of artificial intelligence and will emphasize how AI is poised to improve patient care and streamline dental practice operations. Figure 4 presents a workflow for processing and segmenting Cone Beam Computed Tomography (CBCT) data, ultimately for patient presentation. Raw CBCT data is received, which is then processed by a radiology technician or healthcare professional using software tools such as Anatomage Invivo 6.5 and Diagnocat. Utilizing this software, a report detailing soft tissue characteristics, skeletal and dental structures, as well as potential facial asymmetries is generated automatically. Additionally, the analysis tracks the inferior alveolar nerve pathway and examines airway passages. The process may also include orthopantomograms (OPGs) and cephalometric analyses. As a result of this analysis, a conclusion is reached, which is then conveyed to the patient. An example of how advanced digital tools can enhance patient care by providing understandable diagnostic insights. CBCT data processing workflows can be enhanced by ChatGPT by automating the generation of interpretive reports from CBCT data. The software is capable of synthesizing complex diagnostic data into easy-to-understand summaries. ChatGPT facilitates improved communication between patients by converting technical jargon from advanced software analyses into clear, concise language. Through this process, patients gain a better understanding of the implications of their CBCT scans, such as soft tissue parameters, skeletal structures, and any identified asymmetries or obstructions to their airways.

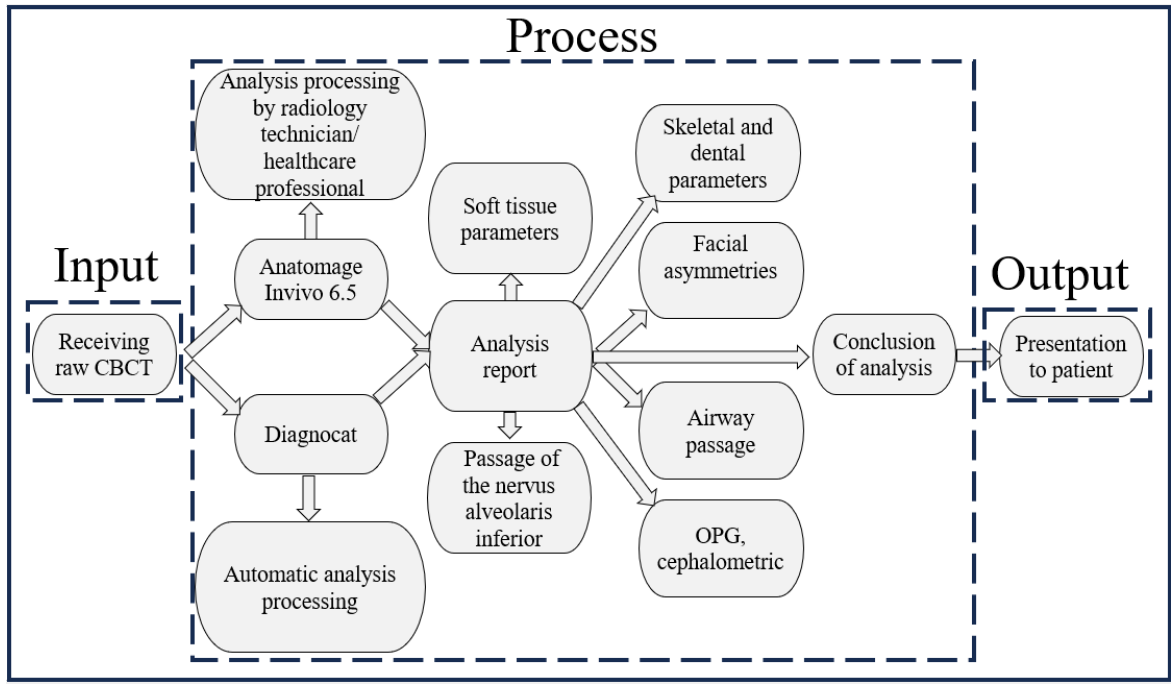

**Figure 4**: Workflow for Processing and Segmenting CBCT Data for Patient Presentation

The ChatGPT application can also provide personalized explanations of the treatment implications based on the diagnostic conclusions, thus bridging the gap between technical medical findings and patient education. As a result, patients are more informed and engaged, which is crucial for successful healthcare outcomes. A study conducted by Shan et al. [80] underscores that AI serves as a complementary tool for dental professionals rather than a substitute. It also emphasizes the current advantages of AI in improving diagnostic accuracy, aiding in treatment planning, and predicting outcomes. Moreover, they mention the challenges associated with AI's broader adoption, such as data management, algorithm transparency, computational demands, and ethical implications. The authors suggest a collaborative approach to integrate AI into dentistry, suggesting that AI can improve efficiency and outcomes, but cautioning that challenges such as methodological limitations, data scarcity, and ethical concerns must be addressed, with human oversight and evidence-based practices being crucial to maintaining trust in AI applications in dentistry. The review by [81] also highlights the significance of 'explainable AI' to demystify AI processes for users and the necessity of updating dental education to include digital proficiency to keep up with the pace of technological advancement.

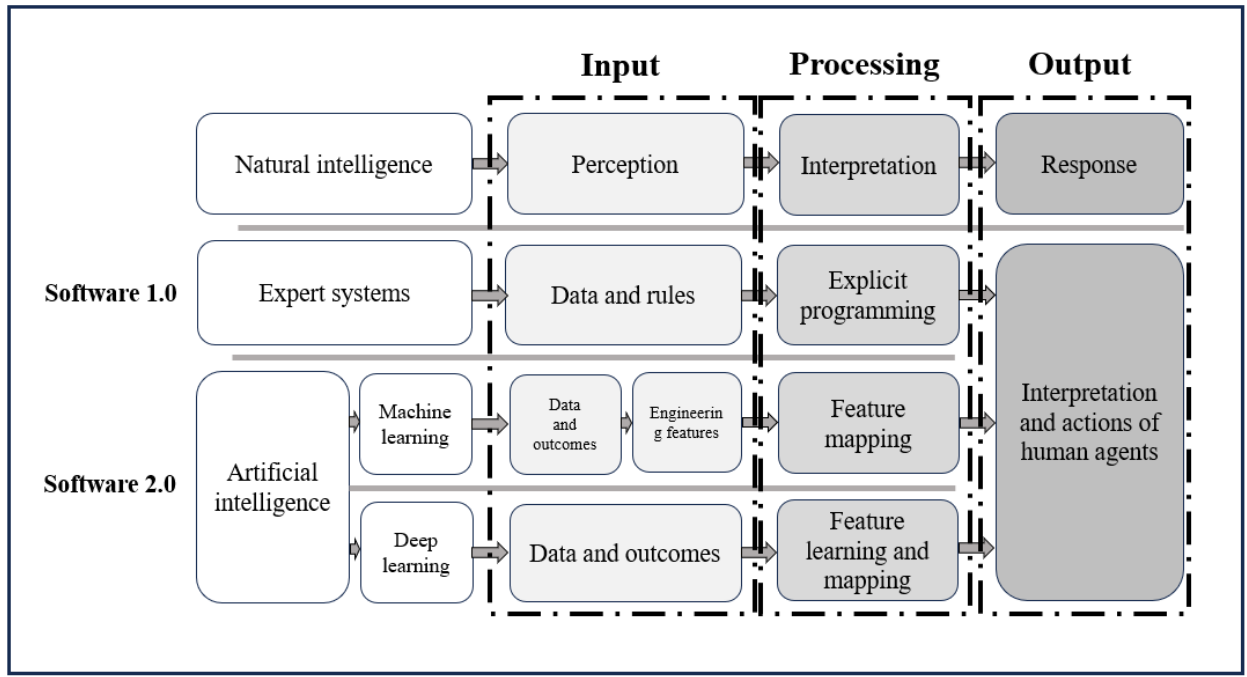

**Figure 5:** Natural VS Computer Intelligence

Li et al. [82] discuss GPT-4's potential in neurosurgery, including its contribution to brain-computer interfaces and understanding neural behaviors, offering insights into therapeutic approaches for neurological conditions. Figure 5 in the study, illustrates the progression from early expert systems to contemporary AI in software development, emphasizing their use in clinical dental practice and surgery. Traditional intelligence follows a linear path from perception to interpretation and then to response, mirroring human cognitive function. 'Software 1.0' represents early expert systems that mimic human cognition by applying predefined data and rules through explicit programming, but these systems require human interpretation of their outputs. 'Software 2.0', encompassing machine learning and deep learning, represents a more advanced phase of AI where systems learn from data outcomes. In machine learning, experts design features to guide algorithms, which are then mapped to outputs [83]. Deep learning, however, automates this by learning from raw data, allowing for more complex data processing and outputs that more closely mimic autonomous human responses. Further discussions on AI's rapid growth in healthcare attribute advancements to increased computational power and data availability [84]. This section intends to provide surgeons with the knowledge to utilize AI's potential, especially in oral and maxillofacial surgery. It notes a surge in AI research within OMFS, particularly the use of convolutional neural networks from machine learning for image-based diagnostics.

### 5-1- AI applications in dentistry for diagnosis, and treatment

Artificial Intelligence (AI) applications in dentistry are transforming the landscape of diagnosis and treatment, offering unprecedented precision, efficiency, and personalized care options for patients. According to He et al. [85], ChatGPT and GPT-4 are beneficial for spinal surgeons, and further advancements in ChatGPT should contribute to surgical planning and patient communication. They caution, however, that these tools should be used cautiously and under close supervision since they lack the nuanced clinical expertise of human physicians. Despite their ability to process large datasets, they are still auxiliary to a physician's judgment. There have been concerns raised about the potential for errors with GPT-4 and the challenges associated with assessing its performance by Nori et al. [86], emphasizing the importance of utilizing new technologies responsibly in order to maximize their benefits and minimize their risks. The versatility of GPT-4 can be seen in its applications in clinical medicine, including as virtual assistants across surgical disciplines and in structuring radiology reports [87-89]. Balel et al. [90] acknowledge ChatGPT's utility in disseminating patient information in oral and maxillofacial surgery but recommend caution when using it for training. It is recommended that surgeons integrate ChatGPT into their clinical expertise as it becomes more prevalent in medicine. According to Ferres et al. [91], ChatGPT can amalgamate diverse data types, which could lead to benefits for radiology by combining images, textual data, and patient records for tasks such as report generation. Despite certain limitations and errors, ChatGPT holds the potential to revolutionize healthcare. Hiroj Bagde et al. [92] conducted an extensive systematic review and meta-analysis to assess ChatGPT's applicability as a research tool in dental and medical fields. They reviewed 11 descriptive studies across various medical topics, finding ChatGPT's accuracy to vary significantly, from 18.3% to 100%. Despite the meta-analysis showing ChatGPT's relative accuracy to be higher than chance, with an odds ratio of 2.25 and a relative risk of 1.47, considerable heterogeneity among studies was noted. The findings suggest ChatGPT can provide appropriate answers in medicine and dentistry, yet caution towards its reliability is advised, indicating a need for further research to enhance its functionality and dependability in these areas. Another study [93] investigated ChatGPT-4's potential as an intelligent virtual assistant in oral surgery. By evaluating ChatGPT-4's responses to 30 oral surgery-related questions, an expert oral surgeon adjudicated discrepancies, resulting in a 71.7% accuracy rate. This underscores ChatGPT-4's utility as a supplementary tool for clinical decision-making in dentistry yet emphasizes it cannot replace the expertise of a qualified oral surgeon. The study calls for further research to ensure AI technologies like ChatGPT-4 are safely and effectively used in oral surgery and other dental specialties.

Chau et al. [94] evaluated ChatGPT versions 3.5 and 4.0 against dental licensing exams from the US and UK. ChatGPT 4.0 showed improved performance, accurately answering 80.7% and 62.7% of questions from the UK and US exams, respectively, and even passing both exams. This marked improvement over ChatGPT 3.5 highlights the potential of Generative Artificial Intelligence in dental education and professional development, though the study calls for further advancements in GenAI. A study by Rizwan and Sadiq [95] assessed ChatGPT's capability in diagnosing and managing cardiovascular diseases (CVDs) through ten hypothetical clinical cases. With a successful diagnosis rate in eight cases, ChatGPT's responses, verified by cardiologists, aligned with current medical guidelines. Despite its limitations in providing specific treatment plans, ChatGPT was found to be a useful tool for medical professionals, particularly for junior doctors, in formulating diagnosis and treatment strategies, highlighting the importance of complementing AI with expert clinical judgment. Batra et al. [96] explored AI's transformative impact on the dental industry, focusing on its potential to enhance diagnostic and treatment accuracy. The study addresses AI's application in various dental care aspects, including sophisticated imaging, disease detection, and the creation of personalized treatment plans. It also considers ethical and privacy concerns related to AI in dentistry, envisioning a future where AI-integrated dental practices lead to more efficient, patient-centered, and effective oral health outcomes, emphasizing the shift towards leveraging technology to improve the quality of patient care. According to Cheng et al. [97] and Hassam et al., [98] GPT-4 can enhance clinical support for spinal surgeons and joint arthroplasty by creating AI-based virtual assistants. A study by Rao et al. [99] demonstrated ChatGPT's utility in radiologic decision-making and clinical support, while other research has demonstrated its effectiveness in providing dental hygiene and health information to patients. The results of these studies demonstrate that ChatGPT and GPT-4 are becoming increasingly important as supportive tools in healthcare, enhancing professional practice and patient care. There are several studies that demonstrate the benefit of artificial intelligence in diagnostics, therapeutic decisions, surgical planning, and prognosis [111]. AI is capable of enhancing learning, classification, prediction, and detection, thus potentially reducing human error and augmenting the skills of clinicians. Furthermore, they highlighted the need for AI algorithms to undergo rigorous clinical validation and ethical scrutiny regarding data protection [112-114].

**Table 3:** A survey of AI applications in dentistry

| AUTHORES | Year | Aim | Method | Result |
|---|---|---|---|---|
| Alzaid et al.[100] | 2023 | To provide an overview of AI applications in various dental specialties. | Exhaustive literature search on AI applications in dentistry. | Highlighted AI's role in improving disease diagnosis and treatment planning, though AI cannot replace dentists. |
| Tiwari et al.[101] | 2023 | Investigate ChatGPT's applications in public dental health for research, education, and clinical practice. | Systematic literature review on ChatGPT's implications in public health dentistry. | ChatGPT aids in scientific writing and research, with potential benefits and risks noted. |
| Khurana, and Vaddi [102] | 2023 | Explore the limitations and potential applications of ChatGPT in academic oral and maxillofacial radiology (OMFR). | Editorial overview based on authors' experiences and relevant literature. | Identified valuable applications in education and limitations in image-based question answering and authorship validity. |
| Alhaidry et al.[103] | 2023 | Discuss the use of ChatGPT in dentistry for diagnoses, disease risk assessment, and other applications. | Review of literature on ChatGPT's use in dentistry. | ChatGPT benefits include anomaly detection and workload reduction, but with notable risks and limitations. |
| Bagheri and Varzaneh[104] | 2023 | Review applications of ChatGPT and GPT4 in biology, medical and dental studies, and health care, along with concerns about its use. | Article review focusing on ChatGPT and GPT-4's contributions and limitations in science and healthcare. | Highlighted ChatGPT's impact on various fields despite limitations and raised concerns on potential misuse. |
| Zhou et al. [105] | 2023 | Examine how AI can transform dentistry, improve patient outcomes, and streamline tasks, with a focus on ethical, legal, and regulatory implications. | Analysis of AI applications in dentistry, particularly in diagnosis and treatment planning. | AI improves diagnosis accuracy and operational efficiency but requires careful management of over-reliance and ethics. |
| Kavadella et al.[106] | 2024 | Evaluate ChatGPT's implementation in dental education quantitatively and qualitatively. | Mixed methods study with 77 dental students using ChatGPT for a learning assignment. | ChatGPT group performed better in knowledge exams; students recognized its benefits and limitations. |
| Kamath et al.[107] | 2024 | Highlight AI's role in dentistry and predict its future impacts. | Narrative review of 59 papers from Google Scholar and PubMed. | AI useful in various dental phases; future advancements anticipated in dental tools and education. |
| Büttner et al.[108] | 2023 | Summarize NLP applications and limitations in dentistry. | Narrative review. | NLP has potential for applications in dentistry, though challenges remain in its broader adoption. |
| Younis et al.[109] | 2024 | Explore AI's transformative potential in healthcare, including dentistry. | Systematic literature review and meta-analysis of 82 papers. | ChatGPT and AI tools show promise in medicine and healthcare for various applications, with ongoing challenges. |

| Ali et al.[110] | 2023 | Evaluate ChatGPT's performance in dental education assessments. | Exploratory study on ChatGPT's accuracy in dental curricula assessments. | ChatGPT provided accurate responses to most assessments, indicating potential to revolutionize virtual learning and assessments. |
|---|---|---|---|---|
| Bragazzi et al.[111] | 2023 | Assess ChatGPT's diagnostic accuracy in endodontics. | Analysis of 70 peri-apical X-rays by ChatGPT. | ChatGPT showed limited clinical usability in current form, with a need for improvements in dental diagnostics accuracy. |

The medical field often emphasizes algorithmic thinking, but healthcare providers sometimes resort to heuristic decision-making due to heavy workloads and cognitive demands, which can lead to errors due to factors such as experience, emotional state, fatigue, and personal characteristics. AI is rapidly being developed in healthcare and is praised for its ability to support medical decision-making processes and minimize cognitive biases and errors. According to the studies, AI methodologies may be beneficial at all stages of patient care, from initial screening to recovery post-surgery [115].

## 6- An Exploration of the Capabilities of LLMs in Dentistry

LLMs can automate the comprehension of documents and enable the analysis of treatment plans, leveraging the advantages of extensive pretraining. Additionally, exposure to billions of documents helps LLMs develop natural language reasoning (NLR) capabilities for context understanding. This NLR capability can enhance the efficiency of dental practitioners in crafting personalized treatment plans based on patients' backgrounds. For example, NLR algorithms can scrutinize patterns of adverse drug reactions (ADRs) associated with various dental procedures and medications. Understanding common ADRs linked to drugs allows dentists to adjust their treatment plans, reducing the likelihood of side effects such as gum bleeding and severe conditions like bisphosphonate-related osteonecrosis. Alexander Fuchs et al. [116] evaluated the effectiveness of ChatGPT versions 3 and 4 by employing questions from the European Examination in Allergy and Clinical Immunology (EEAACI) and the Swiss Federal Licensing Examination in Dental Medicine (SFLEDM). They explored the effect of priming on enhancing ChatGPT's responses. ChatGPT 4 notably outperformed version 3 in all tests, with priming improving its performance, especially for SFLEDM questions. This study illustrates the rapid advancement in large language model technology and the potential benefits of priming yet cautions against the uncritical application of such models in healthcare due to inherent risks and limitations. A study [117] at Meharry Medical College examined ChatGPT's integration into the dental curriculum, highlighting both its potential and challenges in enhancing dental education. The research analyzed the chatbot's responses to various queries using course materials, showcasing ChatGPT's utility in assisting with academic writing and content creation. Despite identifying limitations, the study emphasizes the importance of rigorous evaluation and the combination of AI technologies like ChatGPT in dental education to improve teaching and learning quality.

**Table 4:** An overview of existing medical LLMs and their model development

| Domains | Model Development Archetypes | Scale of Parameters | Data Magnitude |
|---|---|---|---|
| Pre-training | B-ERT Large | Extensive | Large-scale data |
| | PubM-ERT Large | Medium | Considerable corpus |
| | ClinicalB-ERT | Medium | Broad clinical notes |
| | MedC-ERT | Medium | Ample scientific articles |
| | BioM-ELMo | Small | Large data repository |
| | OphthGPT | Medium | Substantial dialogues |
| | GatorTron Large | Medium | Large token database |
| | GatorTron GP | Small | Moderate token database |
| | MEDTRON | Small | Extensive notes |
| | DoctorLM | Medium | Vast dialogues |
| | BioQuest | Medium | Extensive dialogues |
| | ClinicalGPT | Small | Wide EHRs and QA |
| | QliM-Med | Small | Broad dialogues |
| Medical-domain LMs | CharDoctor | Small | Focused instructions |
| | BenTaa | Small | Specific instructions |
| | HuatoGP | Small | Extensive instructions and dialogues |
| | Baize-healthcare | Medium | Broad dialogues |
| | MedAePac | Small/Medium | Focused medical QA |
| | AlpCure | Small | Specific instructions |
| | Zhiyin | Small | Dedicated instructions |
| | PMCLLM | Medium | Extensive tokens |
| Fine-tuning | CPLLM | Medium | EHRs |
| | Clinical Camel | Small/Medium | Extensive articles and QA |
| | MedLP-M | Medium | Focused medical QA |
| | BioPump | Medium | Extensive tokens |
| | CoeDP | Medium | Extensive thought pieces |
| | CadX | Small | Specific instructions |
| | Dr. Kumo | Small | Zero-shot learning |
| | CharPZ | Small | Zero-shot learning |
| | MedPrompt | Medium | Dedicated prompting |
| | MedPromptM | Medium | Few-shot learning |

A comparative study [28] evaluated the performance of GPT-3.5, GPT-4, and Google Bard using the Japanese National Dentist Examination (JNDE) to assess their clinical application potential in Japan. GPT-4 led in overall correctness, with Bard excelling in essential questions, indicating the distinct capabilities and possible clinical uses of these models in dentistry. The findings suggest a promising future for LLMs in clinical settings, though further validation is needed to ascertain their global utility. Koubaa et al. [118] provided a review of ChatGPT, analyzing its technological advancements and the breadth of research surrounding it. This first extensive literature assessment on ChatGPT outlines future research challenges and directions, offering valuable insights for stakeholders interested in the model's applications, implications, and development prospects. Ana et al. [119] critically appraised studies observing the connection between dentofacial features and dental trauma among Brazilian children and adolescents, focusing on the handling of confounding factors. Despite some acknowledgment of biases, the review found a general lack of thorough consideration for confounders in these studies, underscoring the difficulty in establishing causality from such research. This highlights the need for more rigorous methodological approaches in observational studies within the field. The goal of Supervised Fine-Tuning (SFT) is to make use of high-quality medical corpora, such as knowledge graphs [122], physician-patient discussions [120], and medical question-answering [121]. In order to further pre-train the generic LLMs with the same training objectives, such as next token prediction, the produced SFT data acts as a continuation of the pre-training data. Through the additional pre-training period offered by SFT, general LLMs can become more knowledgeable about medicine and more in line with the medical field, enabling them to become specialized medical LLMs. Supervised fine-tuning (SFT), Instruction fine-tuning (IFT), and parameter-efficient tuning are some of the methods used nowadays for fine-tuning [123]. Table 1 provides an overview of the refined medical LLMs that were produced. The ChatGPT AI model utilized in this study is text-based, hence it cannot directly read radiologic images. Consequently, the final diagnoses were produced using descriptive radiologic results (text data). Future studies could benefit from using AI models for picture segmentation and captioning to generate radiologic findings that are descriptive. ChatGPT can then use these findings as the foundation for further diagnostic conclusions [124]. The process of describing an image's visual information in natural language using a visual understanding system and a language model that can produce coherent, syntactically accurate sentences is known as image captioning [125]. In addition, the newly launched ChatGPT4V now supports the entry of photos in addition to text. Writing radiological reports could undergo more alterations as a result of all these AI models.

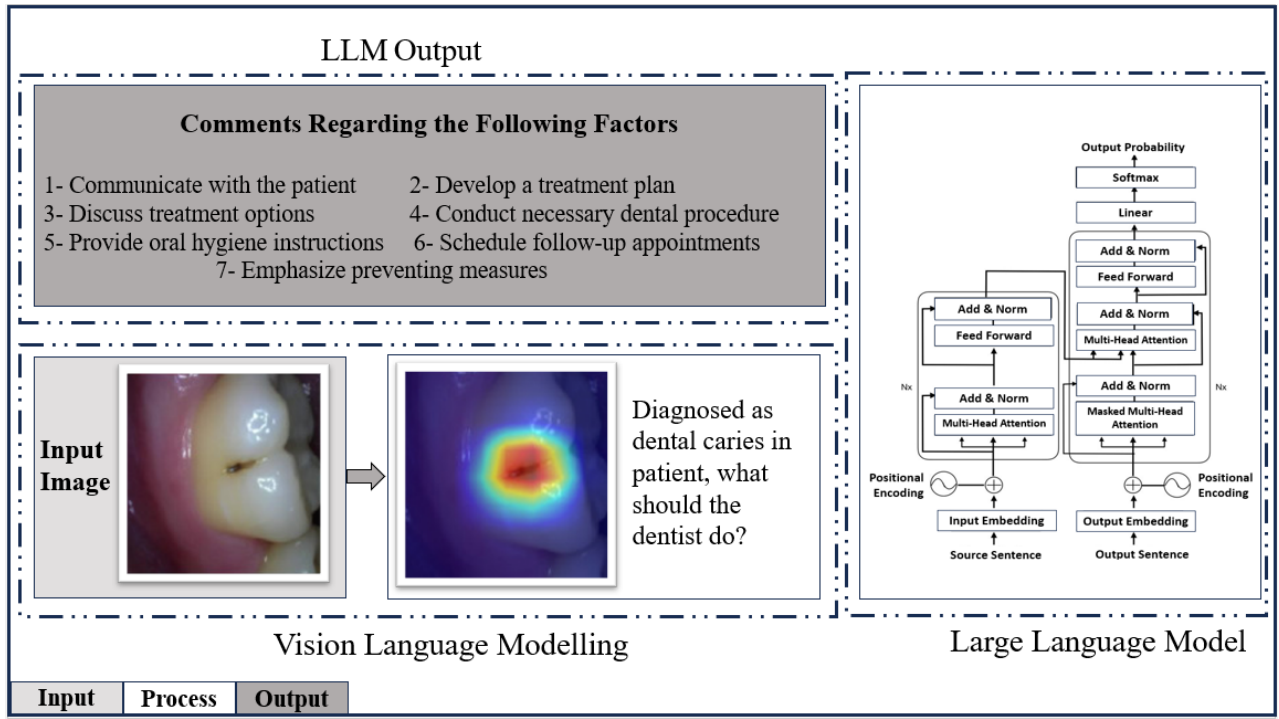

**Figure 6:** process of Large Language Model (LLM) for dental care

Figure 6 illustrates a workflow in which an LLM, integrated with vision language modeling capabilities, processes an input image of a dental condition such as caries and generates a multifaceted output tailored for dental practices. The LLM analyzes the input image to understand the dental issue and generates an action plan for the dentist to follow, illustrating how artificial intelligence can assist with clinical decision-making. Figure 6 illustrates how an LLM can streamline this analysis [32]. By reviewing patient histories, NLR can also identify comorbidities, monitor drug safety, and enhance patient education through natural language generation-based medical documentation utilizes high-quality medical corpora, including physician-patient dialogues [126], medical Q&A [127], and knowledge graphs [128,129], to continue the pretraining of general LLMs with the same objectives, such as predicting the next word or phrase. This additional pretraining phase enables the LLMs to acquire extensive medical knowledge and become attuned to the medical field, effectively transforming them into specialized medical LLMs. Fine-tuning techniques like SFT, Instruction Fine-Tuning (IFT), and Parameter-Efficient Tuning refine these models further. The findings suggest that dentists could be supported in providing thorough, informed care by a sophisticated artificial intelligence system.

# 7- Conclusion

AI has transformed our engagement with technology, with progress in AI algorithms and Large Language Models (LLM) paving the way for the emergence of Natural Generative Language (NGL) systems like ChatGPT. To enhance the assessment of AI-generated information in dental, the paper suggests that future studies should involve a broader and more diverse group of dentists from multiple universities. Additionally, comparing the outputs of ChatGPT with those of dental students and postgraduates could provide additional insight into the limitations and capabilities of AI in this field. Even though ChatGPT is acknowledged as a promising technology, it should be used with caution and should not be substituted for human intellectual labor at this time. While the study highlights ChatGPT's promise for patient and parent education in dental diagnostics and as an initial research aid, it also advocates for advancements in artificial intelligence to guarantee the precision and depth of its responses. In analyzing ChatGPT's role, the benefits and drawbacks of this AI-driven application are discussed. According to other research on artificial intelligence in public health dentistry, AI-driven digital dental assistants may be able to perform better than humans when it comes to scheduling, risk assessment, and treatment planning. Through active data provision and self-management, artificial intelligence may also enhance patient engagement in their healthcare. Although AI integration is beneficial, the paper concludes that a dentist's indispensable role remains since clinical practice encompasses more than diagnosis - it includes providing personalized treatment. As AI technologies such as ChatGPT continue to evolve, their integration into routine oral surgery practices is expected, but this must be done with caution. To ensure that these technologies are used safely and effectively, ongoing research must be conducted to ensure patient safety and the integrity of medical practice are always prioritized.

## 7-1- Limitation

ChatGPT offers considerable promise, but it is still in the early stages of development. The use of ChatGPT for information comes with ethical considerations, and it cannot be claimed as a doctor. However, it can be a valuable clinical tool. The human intellect remains irreplaceable by ChatGPT. To ensure that the model is accurate and reliable before its application in clinical settings, it is essential to verify its accuracy and reliability. To overcome the limitations of the ChatGPT model, it is necessary to continuously update and enhance it, considering user feedback. For healthcare professionals and patients, it is important to establish clear protocols regarding the appropriate use of ChatGPT.

Furthermore, policies must be in place to protect patient confidentiality, and ethical standards must be established to navigate the complex issues presented by the integration of ChatGPT into healthcare.

**Funding**

The funding sources had no involvement in the study design, collection, analysis, or interpretation of data, writing of the manuscript, or in the decision to submit the manuscript for publication.

**Compliance with ethical standards**

Conflict of interest Authors have no conflict of interest.